\documentclass{article}
\usepackage{colm2024_conference}

\usepackage{booktabs}
\usepackage{graphicx}
\usepackage{enumitem}
\usepackage{wrapfig}
\usepackage{float}
\usepackage{microtype}
\usepackage{amsmath}
\usepackage{amssymb}
\usepackage{colortbl}
\usepackage[utf8]{inputenc}
\usepackage{caption}
\usepackage{subcaption}
\usepackage{xcolor}
\usepackage{setspace}
\usepackage{url}
\usepackage{multirow}
\usepackage{tabularx}
\usepackage{makecell}
\usepackage{xurl}
\usepackage[misc]{ifsym}

\usepackage{hyperref}
\usepackage[capitalize, nameinlink, noabbrev]{cleveref}

\renewcommand{\arraystretch}{1.25}

\title{
ChatUMM: Robust Context Tracking for Conversational Interleaved Generation
}

\author{
    \vspace{-1em}
    \small
    Wenxun Dai$^{1,2}$, 
    Zhiyuan Zhao$^{2}$\thanks{Project lead. ~~~\Letter~Corresponding author.}, 
    Yule Zhong$^{2}$, 
    Yiji Cheng$^{2}$, 
    Jianwei Zhang$^{2}$, \\
    Linqing Wang$^{2}$, 
    Shiyi Zhang$^{1}$, 
    Yunlong Lin$^{2}$, 
    Runze He$^{2}$, 
    Fellix Song$^{2}$, 
    Wayne Zhuang$^{2}$, \\
    Yong Liu$^{1}$, 
    Haoji Zhang$^{1}$, 
    Yansong Tang$^{1}$, 
    Chunyu Wang$^{2\text{\Letter}}$ \\
    $^{1}$\textbf{Tsinghua University} \quad $^{2}$\textbf{Tencent Hunyuan}
    \vspace{-1em}
}

\begin{document}

\maketitle

\begin{abstract}
Unified multimodal models (UMMs) have achieved remarkable progress yet remain constrained by a single-turn interaction paradigm, effectively functioning as solvers for independent requests rather than assistants in continuous dialogue. To bridge this gap, we present \textbf{ChatUMM}. As a conversational unified model, it excels at robust context tracking to sustain interleaved multimodal generation. ChatUMM derives its capabilities from two key innovations: an interleaved multi-turn training strategy that models serialized text-image streams as a continuous conversational flow, and a systematic conversational data synthesis pipeline. This pipeline transforms a diverse set of standard single-turn datasets into fluid dialogues through three progressive stages: constructing basic stateful dialogues, enforcing long-range dependency resolution via ``distractor'' turns with history-dependent query rewriting, and synthesizing naturally interleaved multimodal responses. Extensive evaluations demonstrate that ChatUMM achieves state-of-the-art performance among open-source unified models on visual understanding and instruction-guided editing benchmarks, while maintaining competitive fidelity in text-to-image generation. Notably, ChatUMM exhibits superior robustness in complex multi-turn scenarios, ensuring fluid, context-aware dialogues.
\end{abstract}
\section{Introduction}

\begin{figure}[!ht]
    \vspace{-1em}
    \centering
    \includegraphics[width=\textwidth]{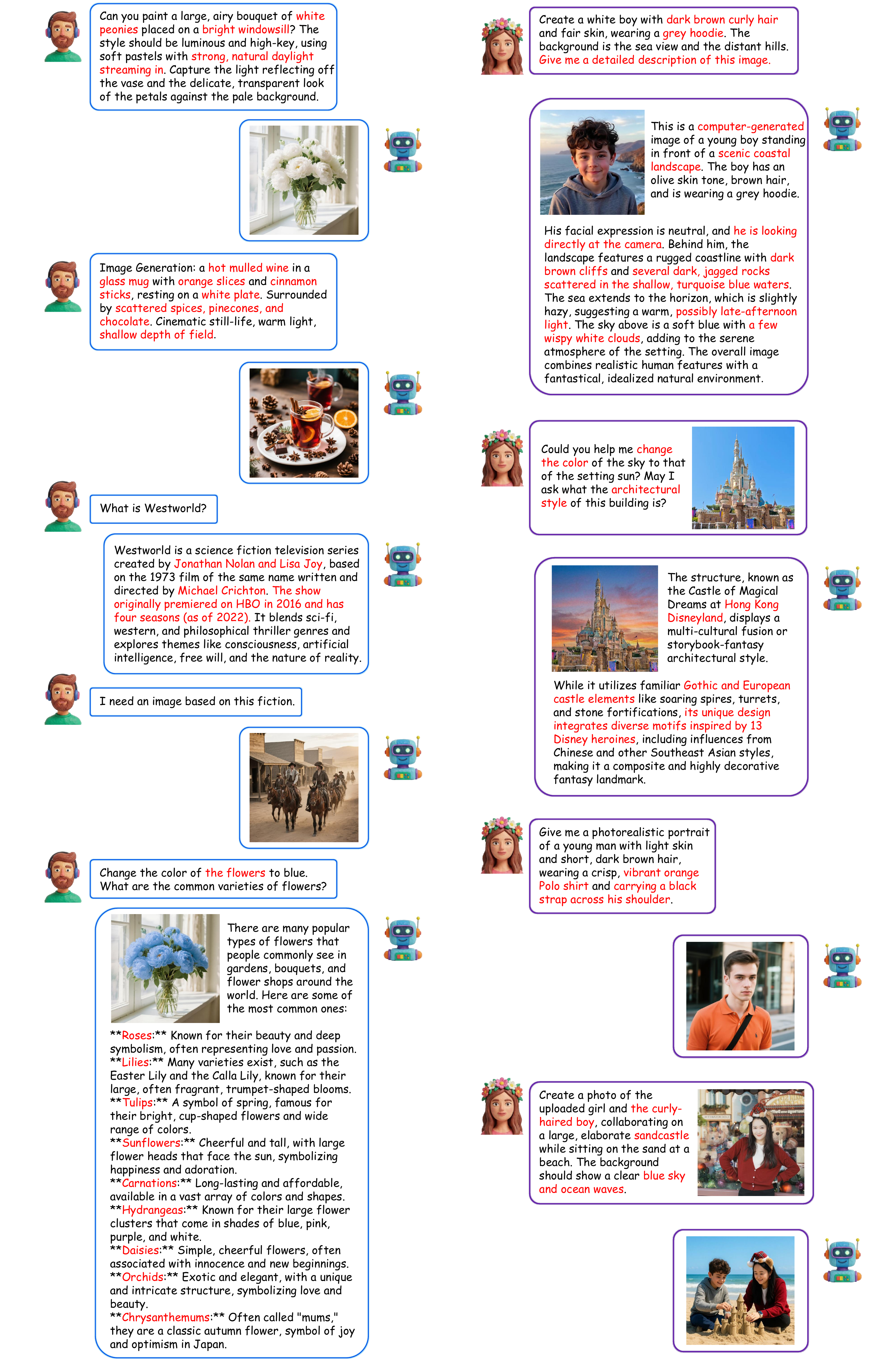}
    \caption{\textbf{Examples of ChatUMM demonstrating diverse conversational capabilities.}}
    \label{figures/teaser}
    \vspace{-1em}
\end{figure}

The emergence of unified multimodal models (UMMs)~\cite{bagel, mmada, metaqueries, mogao, blip3-o, show-o, show-o2, omnigen2, chameleon, janus, janus-pro, transfusion, emu, emu2, emu3, seed-llama, dreamllm}, capable of jointly processing and generating both text and images, represents a significant leap forward for general-purpose multimodal intelligence~\cite{promptenhancer, univg, soc++, kimi-k2, motionlcm, uniworld, lavt, voco}. However, a fundamental gap persists between this vision and the current reality of open-source models~\cite{bagel, show-o2, emu3, janus-pro, transfusion}. These models primarily operate within a single-turn paradigm, functioning as versatile solvers for handling independent requests rather than as assistants in a stateful, evolving conversation. While leading commercial multimodal systems like GPT-4o~\cite{gpt4o} and Doubao~\cite{doubao} circumvent this by employing complex agent-based architectures to coordinate single-task experts, this approach remains a workaround that may inhibit holistic reasoning. In reality, genuine creative processes and complex problem-solving are rarely instant; they unfold as a continuous dialogue where context is built, refined, and referenced over multiple turns. For instance, a user might generate an initial image, follow up with text-based questions, and then refer back to the first image for a history-dependent edit. Such scenarios require a model to move beyond single-task execution and master three critical conversational competencies: \textbf{persistent context tracking, robust long-range dependency resolution, and accurate intent disambiguation across a potentially noisy history}.

To address these challenges, we introduce \textbf{ChatUMM}. As a conversational unified model, it derives its capabilities from the synergy of two key innovations. First, our interleaved multi-turn training strategy (\cref{Methodology}) models serialized text-image streams as a continuous conversational flow, seamlessly accommodating arbitrary forms of multimodal interactions within a unified context. Second, our systematic data synthesis pipeline (\cref{Data Pipeline}) addresses the scarcity of high-quality conversational training data by leveraging atomic, LLM-powered operations to programmatically construct diverse, stateful dialogues. 
This pipeline transforms standard single-turn datasets (e.g., image editing, subject-driven generation) into fluid dialogues through three progressive stages. First, \textit{basic multi-turn construction} converts single-turn samples into basic stateful dialogues. Second, \textit{independent single-turn insertion} injects unrelated ``distractor'' turns and utilizes history-dependent query rewriting, thereby teaching the model to retrieve context across a noisy history. Finally, the \textit{interleaved output generation} stage synthesizes naturally interleaved multimodal responses, where generated images are fluidly integrated with coherent textual feedback. By leveraging this comprehensive dataset with our interleaved multi-turn training strategy, ChatUMM masters robust context tracking, delivering the fluid, context-aware dialogues visualized in \cref{figures/teaser}.

Our primary contributions are as follows:

\begin{itemize}[leftmargin=2em,itemsep=1pt,topsep=0pt,partopsep=0pt,parsep=2pt]

\item We present ChatUMM, a conversational unified multimodal model for interleaved generation. By modeling serialized text-image streams as a continuous conversational flow, it excels at persistent context tracking and accurate intent disambiguation across noisy dialogue history.

\item We propose a systematic data synthesis pipeline that addresses the scarcity of conversational training data. This pipeline leverages LLMs to progressively construct stateful, interleaved multimodal dialogues that enforce robust long-range dependency resolution.

\item Extensive evaluations demonstrate that ChatUMM achieves state-of-the-art performance on visual understanding and instruction-guided editing benchmarks while maintaining competitive fidelity in text-to-image generation. Crucially, it excels in complex multi-turn scenarios, enabling robust and fluid conversational interactions.

\end{itemize}

\section{Methodology}
\label{Methodology}

\begin{figure}[ht!]
    \vspace{-1em}
    \centering
    \includegraphics[width=\textwidth]{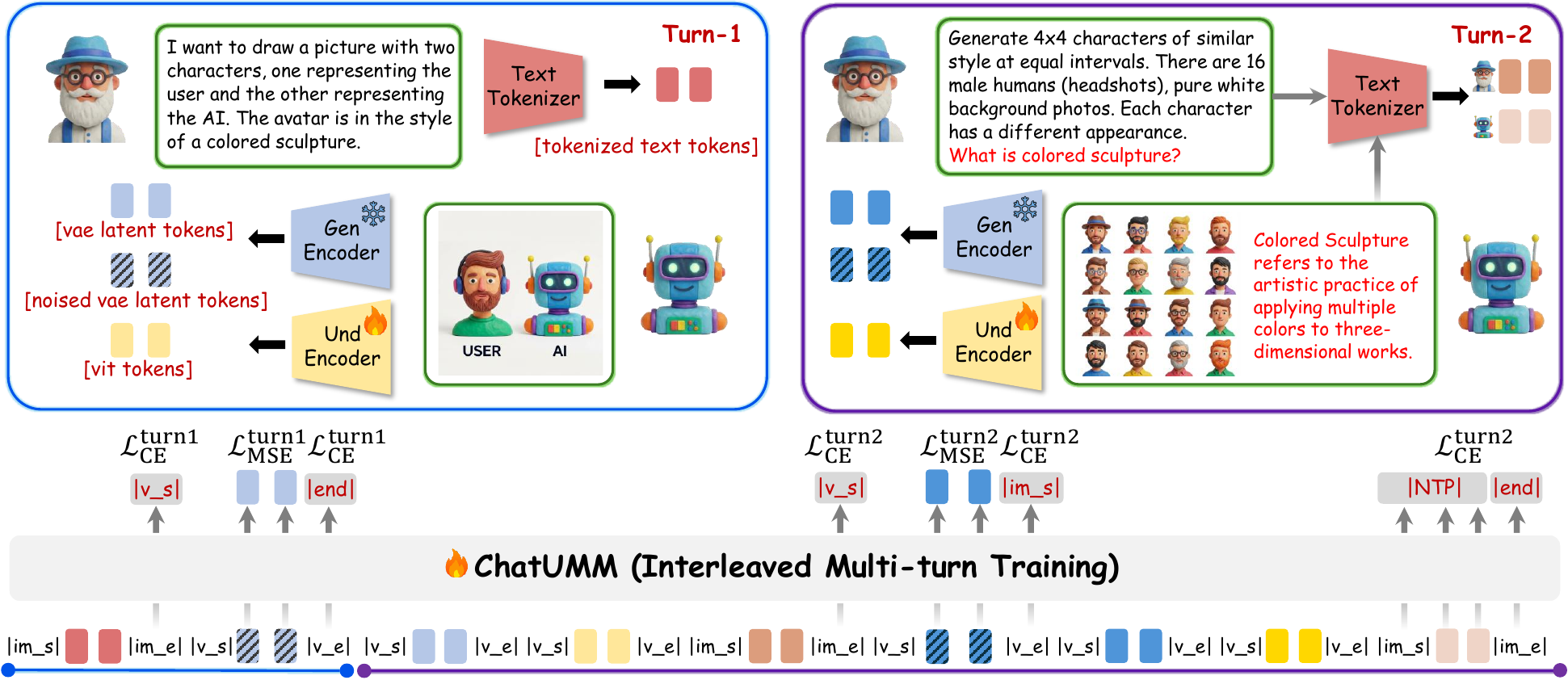}
    \vspace{-1.5em}
    \caption{\textbf{Overview of ChatUMM (Interleaved Multi-turn Training).} The model processes a serialized stream of interleaved text and image tokens using special tokens as structural delimiters. \texttt{|im\_s|} and \texttt{|im\_e|} encapsulate textual segments, while \texttt{|v\_s|} and \texttt{|v\_e|} enclose visual content. Crucially, specific token transitions serve as explicit intent signals: predicting \texttt{|v\_s|} immediately following \texttt{|im\_e|} triggers image generation, whereas predicting \texttt{|im\_s|} after \texttt{|v\_e|} instructs the model to initiate text generation. The \texttt{|end|} token marks the completion of a turn, and \texttt{|NTP|} represents the text tokens generated via next token prediction. For visual generation, the image of the current turn (e.g., Turn-1 and Turn-2) is processed as noised VAE latents (striped blue), supervised by the flow matching loss $\mathcal{L}_{\text{MSE}}$. To retrieve context from the dialogue history (e.g., Turn-1 referenced during Turn-2), the model attends to historical text tokens, clean VAE latents (solid blue), and ViT tokens (yellow). Concurrently, the standard cross-entropy loss ($\mathcal{L}_{\text{CE}}$) is applied to \texttt{|NTP|} and special tokens to supervise text generation and intent prediction.}
    \label{figures/method}
\end{figure}

We introduce ChatUMM, which models multimodal dialogue as a serialized stream to overcome the limitations of single-turn interactions. Unlike models that process independent requests, ChatUMM maintains a continuous conversational flow by integrating interleaved text and image tokens into a unified context. This approach enables fluid, context-aware dialogues, achieved through a unified architecture and a specialized interleaved multi-turn training strategy.

\textbf{Unified Model Architecture.}
Following the architectural design of BAGEL~\cite{bagel}, we employ a decoder-only transformer~\cite{qwen2.5} that incorporates the Mixture-of-Transformers (MoT) mechanism. This structure unifies multimodal understanding and generation by decoupling their conflicting optimization objectives while maintaining a shared context. We adopt a shared self-attention mechanism across all layers to model cross-modal interactions, while utilizing selective activation of modality-specific Feed-Forward Networks (FFNs). Specifically, tokens are hard-routed to either a ``Generation Expert'' (for VAE latents) or an ``Understanding Expert'' (for text and ViT tokens). To address the divergent representational needs of visual perception and generation, the model integrates two separate visual encoders targeting different feature levels: a Vision Transformer (ViT)~\cite{siglip} to capture high-level semantic features for understanding, and a Variational Autoencoder (VAE)~\cite{flux} to preserve low-level visual details in latent representations for generation.

\textbf{Interleaved Multi-turn Training.}
To enable fluid conversational capabilities as shown in~\cref{figures/method}, we introduce an interleaved multi-turn training strategy that models the dialogue as a serialized, continuous flow. The model processes a unified stream of interleaved text and image tokens, employing special tokens as structural delimiters: \texttt{|im\_s|} and \texttt{|im\_e|} encapsulate textual segments, while \texttt{|v\_s|} and \texttt{|v\_e|} enclose visual content. Beyond mere delimitation, these tokens function as control signals for modality switching: the prediction of \texttt{|v\_s|} succeeding \texttt{|im\_e|} initiates image synthesis, while the transition from \texttt{|v\_e|} to \texttt{|im\_s|} activates text generation. To effectively model dependencies across this mixed-modal stream, we employ the Generalized Causal Attention mechanism~\cite{bagel}. Specifically, this mechanism enforces a global causal dependency where both text generation and image synthesis attend to the full interleaved dialogue history. Within this history, the model retrieves context exclusively from text segments, ViT tokens, and clean VAE latents, while explicitly masking out the noised VAE latents of previous turns. This ensures that the current generation is conditioned solely on the noise-free content of the conversation.

\textbf{Optimization Objectives.}
The model is jointly optimized using two objectives. We apply the standard cross-entropy loss ($\mathcal{L}_{\text{CE}}$) for text generation and visual understanding (including special tokens). For visual generation, we employ a Flow Matching~\cite{flow} objective ($\mathcal{L}_{\text{MSE}}$) based on Rectified Flow~\cite{sd3}, which trains the model to predict the velocity field transporting noise to data, ensuring high-fidelity image synthesis within the interleaved sequence.

\section{Data Pipeline}
\label{Data Pipeline}

\begin{figure}[t!]
    \vspace{-1.5em}
    \centering
    \includegraphics[width=\textwidth]{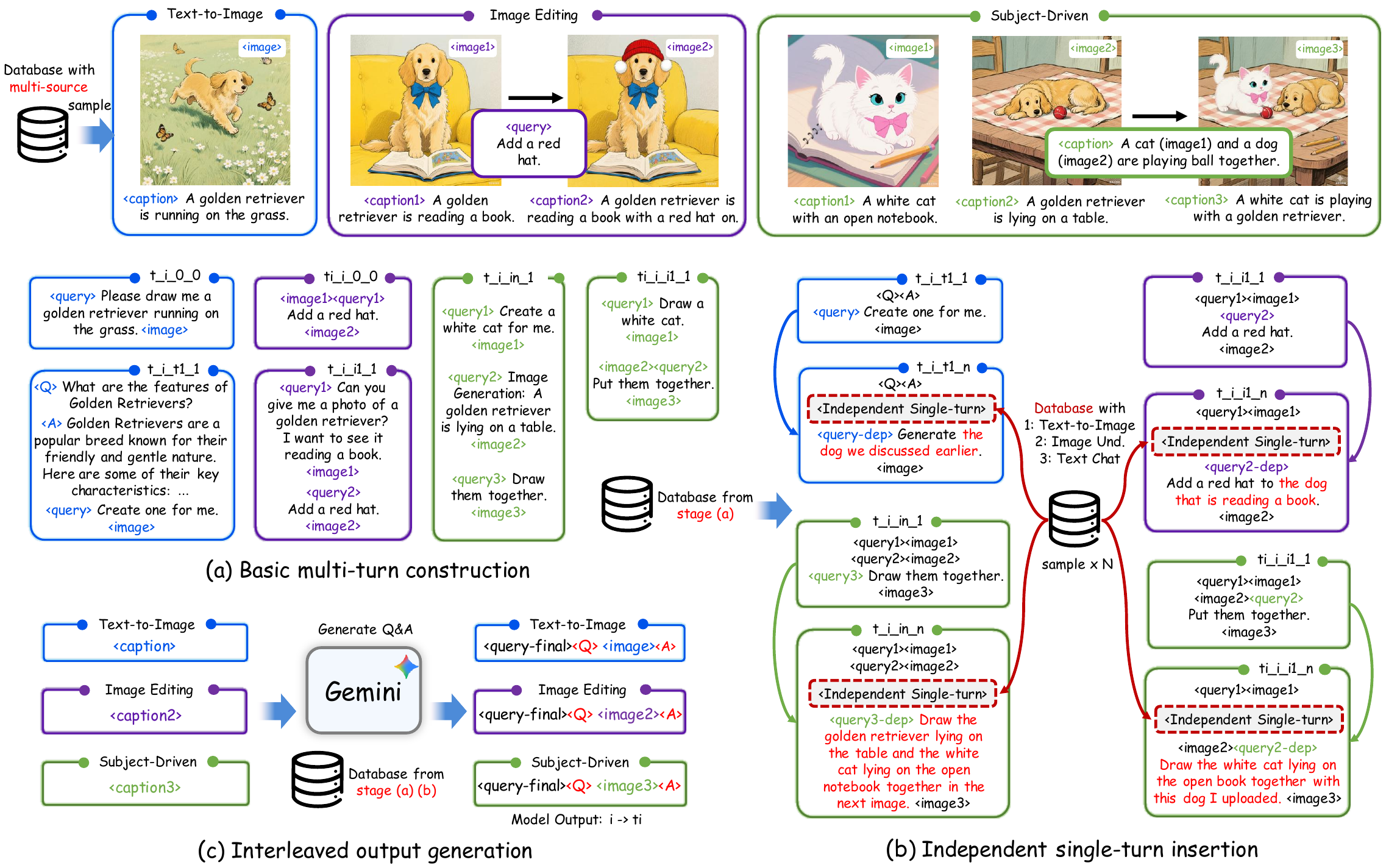}
    \vspace{-1.5em}
    \caption{\textbf{Overview of our data synthesis pipeline.} We transform standard single-turn datasets into fluid, stateful dialogues through three progressive stages. \textbf{(a) Basic multi-turn construction}: We leverage atomic LLM-powered operations to convert single-turn samples (e.g., text-to-image, image editing) into basic stateful dialogues. \textbf{(b) Independent single-turn insertion}: To enforce long-range dependency resolution, unrelated ``distractor'' turns are inserted into the flow. The subsequent query is rewritten to be explicitly history-dependent (\texttt{<query-dep>}), teaching the model to maintain robust context tracking across a noisy history. \textbf{(c) Interleaved output generation}: We evolve the output modality to sustain interleaved multimodal generation. An LLM (e.g., Gemini 2.5 Pro) generates a relevant Q\&A pair based on the image of the final turn, transforming the interaction into a continuous text-image stream (i.e., User: \texttt{<query-final><Q>}, Assistant: \texttt{<image><A>}).}
    \label{figures/data_pipeline}
    \vspace{-1.5em}
\end{figure}

To address the scarcity of high-quality, multi-turn training data, we propose an automated synthesis pipeline (\cref{figures/data_pipeline}) that transforms single-turn datasets into complex conversational flows. We first establish a taxonomy in \cref{Taxonomy of Conversational Data} to categorize data based on input/output modalities, historical dependency modalities, and dependency depth, followed by a library of atomic, LLM-powered operations in \cref{LLM-powered Atomic Operations}. Finally, we employ these operations across three progressive stages: constructing basic stateful dialogues (\cref{Stage (a): Basic Multi-turn Construction}), inducing long-range robustness via independent single-turn insertion (\cref{Stage (b): Independent Single-turn Insertion}), and enabling interleaved multimodal generation (\cref{Stage (c): Interleaved Output Generation}).

\subsection{Taxonomy of Conversational Data}
\label{Taxonomy of Conversational Data}

To systematically categorize our diverse conversational data, we introduce a formal four-dimensional taxonomy. Each data sample is classified based on the following core attributes:

\begin{enumerate}[leftmargin=*,label=\arabic*.]
    \item \textbf{User Input Modality}: The type of input provided by the user in the final turn. This can be either text-only (\texttt{t}) or text and image (\texttt{ti}). Text-only (\texttt{t}) refers to commands issued entirely through text (e.g., ``Make the sky blue''). In contrast, text-and-image (\texttt{ti}) involves providing both a textual instruction and a reference image. This is crucial for tasks like image editing, e.g., uploading a pet photo to ask, ``Place my dog in a magical forest.''
    
    \item \textbf{Model Output Modality}: The type of output the model generates in the final turn. This can be either image-only (\texttt{i}) or text and image (\texttt{ti}). An image-only output (\texttt{i}) is the standard for most generative tasks, where the model's turn is to produce the requested visual content. In contrast, text-and-image (\texttt{ti}) enables continuous interaction by following the generated image with a relevant text response, seamlessly integrating visual generation into the textual dialogue flow.
    
    \vspace{2em}
    
    \item \textbf{Historical Dependency Modality}: The modality and quantity of historical turns the final turn depends on. This is classified as none (\texttt{0}), text history (\texttt{t1}, \texttt{tn}), or image history (\texttt{i1}, \texttt{in}). Zero dependency (\texttt{0}) denotes a context-free final turn. Text dependency (\texttt{t1}, \texttt{tn}) occurs when the final turn references prior textual dialogue (e.g., ``Draw one for me'' after discussing lions). Image dependency (\texttt{i1}, \texttt{in}) is crucial for stateful visual tasks like editing (e.g., ``Make the hat red'') or composition, requiring the final turn to retrieve and manipulate specific visual history.
    
    \item \textbf{Historical Dependency Depth}: The number of turns separating the final turn from the historical context it relies on, classified as zero (\texttt{0}), immediate (\texttt{1}), or long-range (\texttt{n}). Zero dependency (\texttt{0}) signifies a context-free instruction. A depth of one (\texttt{1}) represents short-range dependency (e.g., editing the immediately preceding image). Crucially, a depth of \texttt{n} ($\texttt{n} > 1$) represents long-range dependency, where the target context is separated by unrelated ``distractor'' turns. Training on this data is essential for robust context tracking, teaching the model to disregard unrelated intermediate information and resolve long-range dependencies.
\end{enumerate}

This pipeline establishes a clear naming convention: \texttt{<input>\_<output>\_<dependency>\_<depth>}. For instance, the identifier \texttt{t\_i\_i1\_1} denotes a text input (\texttt{t}) generating an image (\texttt{i}), conditioned on one historical image (\texttt{i1}) from the immediately preceding turn (depth \texttt{1}).

\subsection{LLM-powered Atomic Operations}
\label{LLM-powered Atomic Operations}

\begin{figure}[ht!]
    \vspace{-1.25em}
    \centering
    \includegraphics[width=0.95\textwidth]{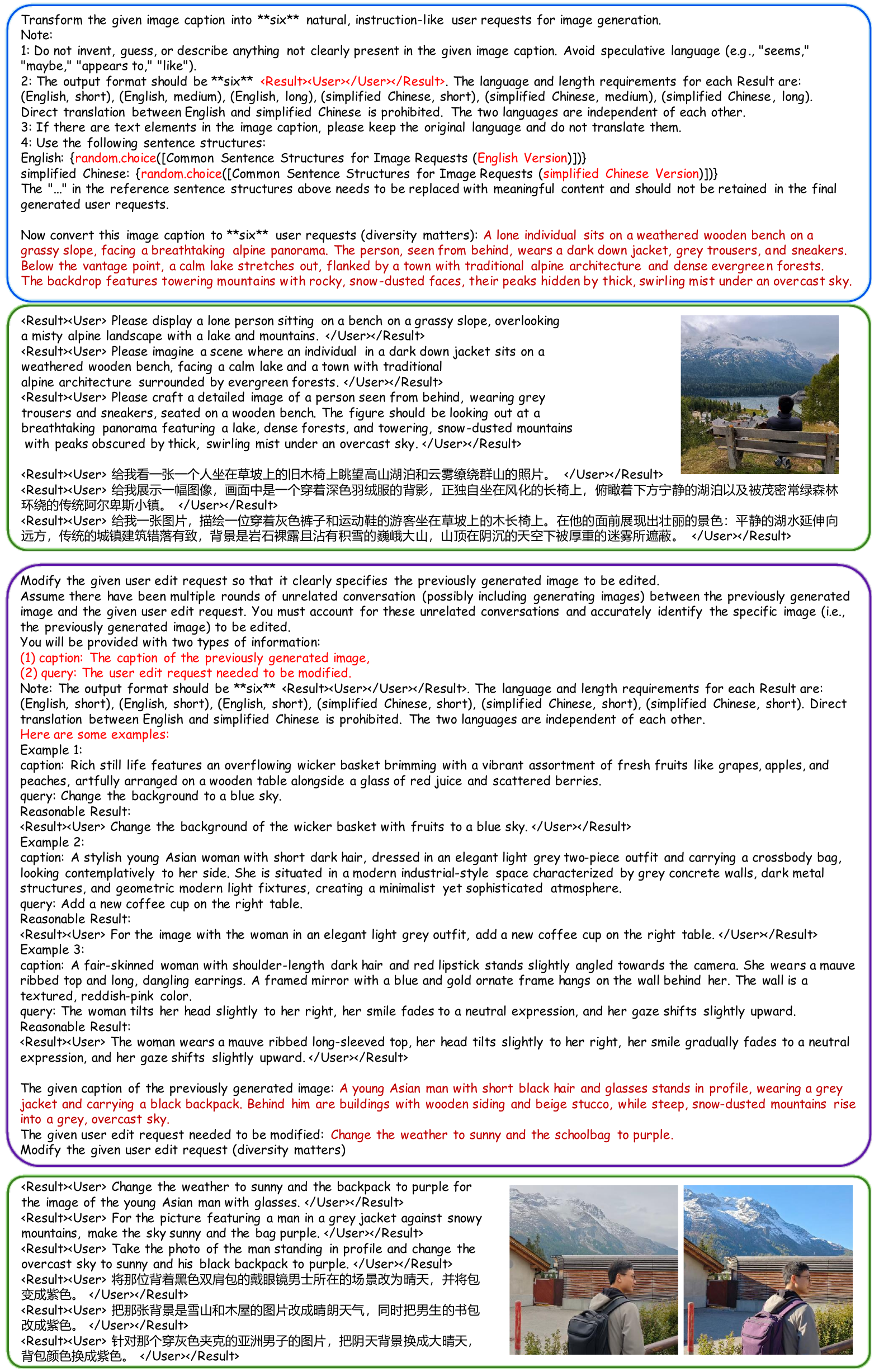}
    \caption{\textbf{Visualization of selected atomic operations.} \textbf{Top:} \texttt{caption2query} converts image captions into user queries. \textbf{Bottom:} \texttt{query2dep\_q} transforms user queries into specific, explicit instructions that clearly identify the target image or subject.}
    \label{figures/prompt}
    \vspace{-1.25em}
\end{figure}

To streamline the data synthesis pipeline, we design a suite of atomic, LLM-powered operations serving as modular building blocks for conversational logic. These operations accept structured text inputs (e.g., captions, queries) and synthesize conversational components, transforming single-turn data into fluid, context-aware dialogues. We categorize them into three groups corresponding to our pipeline stages: basic dialogue construction, long-range dependency injection, and interleaved output generation, as detailed in \cref{tab:prompts}. Selected operations are visualized in \cref{figures/prompt}.

\begin{table}[htbp]

\centering
\caption{\textbf{Overview of LLM-powered atomic operations.} These operations serve as modular building blocks within our data synthesis pipeline, each designed to perform specific text-based transformations (e.g., on captions, queries) to construct fluid, stateful dialogues. They are categorized by the pipeline stage in which they are primarily employed.}

\label{tab:prompts}

\renewcommand{\arraystretch}{3}

\begin{tabularx}{\textwidth}{@{}lX@{}}
\toprule

\rowcolor{gray!20}
\textbf{Atomic Operation} & \textbf{Description} \\
\midrule

\multicolumn{2}{l}{\textit{\textbf{Stage (a): Basic Multi-turn Construction}}} \\
\texttt{caption2query} & Converts an image caption into a user query for generating the image. \\
\rowcolor{gray!10}
\texttt{caption2QA\_q} & Transforms an image caption into a Q\&A pair followed by a generic user query (e.g., ``Create one for me") to initiate history-dependent generation. \\
\texttt{drive\_hs} & Synthesizes a subject-driven query combining two subjects from the two immediately preceding turns (e.g., ``Draw them together''). \\
\rowcolor{gray!10}
\texttt{drive\_i\_h} & Similar to \texttt{drive\_hs}, but combines the subject in the uploaded image with a subject from the immediately preceding turn. \\
\midrule

\multicolumn{2}{l}{\textit{\textbf{Stage (b): Independent Single-turn Insertion}}} \\
\texttt{query2dep\_q} & Rewrites a user query (e.g., ``Add a red hat'') into a specific, explicit instruction (e.g., ``Add a red hat to the dog that is reading a book'') that resolves the reference to the image or subject to be edited. \\
\rowcolor{gray!10}
\texttt{caption2QA\_q\_dep} & Transforms an image caption into a Q\&A pair followed by a history-dependent user query (e.g., ``Generate the dog we discussed earlier''). \\
\texttt{drive\_hs\_dep} & Synthesizes a subject-driven query explicitly combining subjects from two prior turns separated from the current turn by ``distractor'' turns (e.g., ``Draw the golden retriever lying on the table and the white cat lying on the open notebook together in the next image"). \\
\rowcolor{gray!10}
\texttt{drive\_i\_h\_dep} & Similar to \texttt{drive\_hs\_dep}, but combines the subject in the uploaded image with a subject from a prior turn separated by ``distractor'' turns. \\
\midrule

\multicolumn{2}{l}{\textit{\textbf{Stage (c): Interleaved Output Generation}}} \\
\texttt{Q\_from\_caption} & Generates a relevant general-knowledge question based on an image caption. \\
\rowcolor{gray!10}
\texttt{A\_from\_caption} & Generates a factual answer to the question produced by \texttt{Q\_from\_caption}. \\
\bottomrule
\end{tabularx}

\renewcommand{\arraystretch}{1}

\end{table}

\subsection{Stage (a): Basic Multi-turn Construction}
\label{Stage (a): Basic Multi-turn Construction}

\vspace{-0.5em}

This initial stage transforms standard single-turn datasets into elementary, stateful multi-turn dialogues. We source diverse high-quality data from text-to-image, image editing, and subject-driven generation tasks to construct dialogues with a dependency depth of \texttt{0} or \texttt{1}.

\begin{itemize}[leftmargin=2em,itemsep=1pt,topsep=0pt,partopsep=0pt,parsep=2pt]

    \item \textbf{Single-Turn Text-to-Image (\texttt{t\_i\_0\_0})}: Using the \texttt{caption2query} operation, a descriptive image caption (e.g., ``A golden retriever is running on the grass'') is converted into a natural user request for image generation, forming a single-turn interaction with no historical dependency.
    
    \item \textbf{Basic Q\&A-based Generation (\texttt{t\_i\_t1\_1})}: We employ \texttt{caption2QA\_q} to construct a two-turn dialogue. An image caption generates a Q\&A pair (e.g., Q: ``What are the features of Golden Retrievers?'') serving as the first turn context. The subsequent user request (e.g., ``Create one for me'') references this textual history (\texttt{t1}) at a dependency depth of \texttt{1}.

    \item \textbf{Single-Turn Image Editing (\texttt{ti\_i\_0\_0})}: This represents a standard editing task involving both a text query and a reference image, resulting in a modified image with no historical dependency.
    
    \item \textbf{Sequential Image Editing (\texttt{t\_i\_i1\_1})}: A standard single-step editing task is split into a two-turn dialogue. First, the user requests the original image (e.g., ``Can you give me a photo of a golden retriever? I want to see it reading a book''). In the second turn, a targeted edit (e.g., ``Add a red hat'') is issued. The model must retrieve the image generated in the first turn (\texttt{i1}) to execute the edit, creating a dependency at a depth of \texttt{1}.
    
    \item \textbf{Basic Subject-Driven Generation (\texttt{t\_i\_in\_1} and \texttt{ti\_i\_i1\_1})}: For subject composition, we synthesize multi-step dialogues. In \texttt{t\_i\_in\_1}, we use \texttt{caption2query} twice to generate distinct subjects in consecutive turns (e.g., a white cat, then a golden retriever), followed by \texttt{drive\_hs} to issue a composition command (e.g., ``Draw them together''). Alternatively, for \texttt{ti\_i\_i1\_1}, we generate a single subject (e.g., a white cat) via \texttt{caption2query}, then employ \texttt{drive\_i\_h} to combine it with a newly uploaded image (e.g., ``Put them together'').
    
\end{itemize}

\subsection{Stage (b): Independent Single-turn Insertion}
\label{Stage (b): Independent Single-turn Insertion}

\vspace{-0.5em}

To equip the model with robust context tracking capabilities across noisy histories, we augment the dialogues from Stage (a) by inserting ``distractor'' turns (text-to-image, image understanding, or text chat). The core innovation lies in applying operations like \texttt{query2dep\_q} to transform the user's original query into a history-dependent instruction. This process elevates the dependency depth from \texttt{1} to \texttt{n}, forcing the model to filter out distractors and resolve long-range dependencies.

\begin{itemize}[leftmargin=2em,itemsep=1pt,topsep=0pt,partopsep=0pt,parsep=2pt]

    \item \textbf{Long-Context Q\&A-based Generation (\texttt{t\_i\_t1\_n})}: Originating from a \texttt{t\_i\_t1\_1} sample, the implicit request (e.g., ``Create one for me'') becomes ambiguous after distractor insertion. We employ \texttt{caption2QA\_q\_dep} to rewrite it into a specific instruction (e.g., ``Generate the dog we discussed earlier'') that explicitly references the textual topic from \texttt{n} turns ago.

    \item \textbf{Long-Context Image Editing (\texttt{t\_i\_i1\_n})}: We insert distractors between the initial image generation and the editing request in a \texttt{t\_i\_i1\_1} dialogue. Using \texttt{query2dep\_q}, the ambiguous query (e.g., ``Add a red hat'') is transformed into a specific, explicit instruction (e.g., ``Add a red hat to the dog that is reading a book''). This forces the model to ignore intermediate distractors and accurately retrieve the specific visual context from a depth of \texttt{n}.

    \item \textbf{Long-Context Subject-Driven Generation (\texttt{t\_i\_in\_n} and \texttt{ti\_i\_i1\_n})}: For subject composition, we insert distractors before the final composition request. For \texttt{t\_i\_in\_n}, \texttt{drive\_hs\_dep} expands a vague command (e.g., ``Draw them together'') into a history-dependent query (e.g., ``Draw the golden retriever lying on the table and the white cat lying on the open notebook together in the next image''), explicitly combining subjects from prior turns separated by distractors. Similarly, for \texttt{ti\_i\_i1\_n}, \texttt{drive\_i\_h\_dep} combines a historical generated image with a user upload (e.g., ``Draw the white cat lying on the open book together with this dog I uploaded'').
    
\end{itemize}

\subsection{Stage (c): Interleaved Output Generation}
\label{Stage (c): Interleaved Output Generation}

\vspace{-0.5em}

The final stage evolves the model's output from a single image (\texttt{i}) to an interleaved text-and-image (\texttt{ti}) format. Universally applied to the final turn of dialogues from Stages (a) and (b), we leverage operations like \texttt{Q\_from\_caption} and \texttt{A\_from\_caption} to generate a Q\&A pair derived from the image of the final turn. The user's final instruction (\texttt{<query-final>}) is augmented with the generated question (\texttt{<query-final><Q>}), training the model to produce a interleaved response: the requested image followed by the textual answer (\texttt{<image><A>}). This stage updates task signatures (e.g., \texttt{t\_i\_0\_0} $\rightarrow$ \texttt{t\_ti\_0\_0}), seamlessly integrating visual generation into the textual dialogue flow.

\section{Experiments}

\subsection{Training Details}

\textbf{Stage 1: Conversational Tuning.} We use the pre-trained BAGEL~\cite{bagel} checkpoint and conduct training for 15k steps, consuming approximately 30 billion tokens. During this stage, we adopt a uniform sampling strategy across all multi-turn data categories, updating all model parameters excluding the frozen VAE. The model is optimized using AdamW~\cite{adamw} with $\beta_1=0.9$, $\beta_2=0.95$, $\epsilon=1.0 \times 10^{-15}$, zero weight decay, and a gradient norm clip of 1.0. The learning rate follows a 200-step warmup phase before remaining constant at $2 \times 10^{-5}$. We assign equal weights (1:1) to the CE and MSE losses. Unlike the original BAGEL training recipe, we disable EMA updates and directly utilize the online model as the final checkpoint. For efficient large-scale training, we leverage FSDP~\cite{fsdp} with the \texttt{HYBRID\_SHARD} strategy (num\_shard=32), utilize Flex Attention~\cite{flexattn}, and employ sequence packing with lengths ranging from 61k to 64k per rank, with the diffusion timestep shift set to 1.0.

\textbf{Stage 2: Annealing.} We extend training for an additional 5k steps (10B tokens) while maintaining identical optimization hyperparameters. To enhance multimodal understanding, we incorporate image understanding data with a sampling ratio of 0.25. Simultaneously, to preserve text-based conversational abilities, we include text-only data with a ratio of 0.15. We reinforce foundational image generation capabilities by setting sampling ratios for single-turn text-to-image and image editing data to 0.5 and 1.0, respectively. Accordingly, we adjust the sampling ratios for all multi-turn tasks to 0.1 to retain the acquired stateful conversational abilities.

\subsection{Evaluation Benchmarks}

For multimodal understanding, we adopt six widely used benchmarks to form a comprehensive evaluation suite spanning perception, cognition, and complex reasoning. Specifically, we evaluate on \textbf{MME}~\cite{mme} and \textbf{MMBench}~\cite{mmbench} for general multimodal capabilities, and \textbf{MM-Vet}~\cite{mmvet} to assess integrated visual-linguistic skills. To probe expert-level and mathematical reasoning, we employ \textbf{MMMU}~\cite{mmmu} and \textbf{MathVista}~\cite{mathvista}, respectively. Additionally, we include \textbf{MMVP}~\cite{mmvp} to specifically evaluate fine-grained visual perception patterns. Overall, these datasets enable a holistic evaluation of our model against state-of-the-art models across diverse multimodal understanding tasks.

For generative capabilities, we assess both text-to-image generation and instruction-guided image editing. We evaluate text-to-image generation performance on the \textbf{GenEval}~\cite{geneval} benchmark, which systematically measures text-to-image alignment by assessing object-level compositional capabilities, such as object recall, spatial relationships, counting, and attribute binding. For image editing, we utilize \textbf{GEdit-Bench}~\cite{step1x} as our primary evaluation suite. Curated from authentic web-scraped user requests, GEdit-Bench mirrors practical editing needs and covers a diverse range of tasks. Following standard protocols, we report performance scores automatically evaluated by GPT-4.1~\cite{gpt4.1} to ensure an objective assessment of the model's ability to follow editing instructions while maintaining visual fidelity.

\subsection{Image Understanding}

\begin{table*}[t!]
\vspace{1em}
\centering
\caption{\textbf{Comparison with state-of-the-arts on visual understanding benchmarks.} MME-S refers to the
summarization of MME-P and MME-C. For MoE models, we report their activated params / total params. $\dagger$:
MetaQuery~\cite{metaqueries} adopts the pre-trained model from Qwen2.5-VL~\cite{qwen2.5-vl} and freezes it during training. ${**}$: Partial results are from by MetaMorph~\cite{metamorph} or MetaQuery~\cite{metaqueries}. ${*}$: We report the results without Chain-of-Thought.}
\label{tab:model_performance_und}
\resizebox{\textwidth}{!}{
\begin{tabular}{llcccccccc}
\toprule
\textbf{Type} & \textbf{Model} & \textbf{\# LLM Params} & \textbf{MME-P}$\uparrow$ & \textbf{MME-S}$\uparrow$ & \textbf{MMBench}$\uparrow$ & \textbf{MMMU}$\uparrow$ & \textbf{MM-Vet}$\uparrow$ & \textbf{MathVista}$\uparrow$ & \textbf{MMVP}$\uparrow$ \\
\midrule
\multirow{13}{*}{\rotatebox{90}{Und. Only}} 
& InternVL2~\cite{internvl2}        & 1.8B         & 1440 & 1877 & 73.2 & 34.3 & 44.6 & 46.4 & 35.3 \\
& InternVL2.5~\cite{internvl2.5}       & 1.8B         & -    & 2138 & 74.7 & 43.6 & 60.8 & 51.3 & -    \\
& Qwen2-VL~\cite{qwen2-vl}          & 1.5B         & -    & 1872 & 74.9 & 41.1 & 49.5 & 43.0 & -    \\
& Qwen2.5-VL~\cite{qwen2.5-vl}         & 3B           & -    & 2157 & 79.1 & 53.1 & 61.8 & 62.3 & -    \\
& BLIP-3~\cite{blip-3}            & 4B           & -    & -    & 76.8 & 41.1 & -    & 39.6 & -    \\
& LLava-OV~\cite{llava-ov}          & 7B           & 1580 & -    & 80.8 & 48.8 & 57.5 & 63.2 & -    \\
& InternVL2~\cite{internvl2}         & 7B           & 1648 & 2210 & 81.7 & 49.3 & 54.2 & 58.3 & 51.3 \\
& InternVL2.5~\cite{internvl2.5}       & 7B           & -    & 2344 & \underline{84.6} & 56.0 & 62.8 & 64.4 & -    \\
& Qwen2-VL~\cite{qwen2-vl}          & 7B           & -    & 2327 & 83.0 & 54.1 & 62.0 & 58.2 & -    \\
& Qwen2.5-VL~\cite{qwen2.5-vl}         & 7B           & -    & 2347 & 83.5 & \textbf{58.6} & \underline{67.1} & 68.2 & -    \\
& Emu3-Chat$^{**}$~\cite{emu3}       & 8B           & 1244 & -    & 58.5 & 31.6 & 37.2 & -    & 36.6 \\
& Kimi-VL~\cite{kimi-vl}           & 2.8B/16B     & -    & -    & -    & \underline{57.0} & 66.7 & 68.7 & -    \\
& DeepSeek-VL2~\cite{deepseek-vl2}      & 4.1B/28B     & -    & -    & -    & 51.1 & 60.0 & 62.8 & -    \\
\midrule
\multirow{15}{*}{\rotatebox{90}{Unified}} 
& Show-o$_{512}$~\cite{show-o}         & 1.3B         & 1097 & -    & -    & 26.7 & -    & -    & -    \\
& Janus~\cite{janus}             & 1.5B         & 1338 & -    & 69.4 & 30.5 & 34.3 & -    & -    \\
& Janus-Pro~\cite{janus-pro}         & 1.5B         & 1444 & -    & 75.5 & 36.3 & 39.8 & -    & -    \\
& BAGEL~\cite{bagel}         & 1.5B MoT     & 1610 & 2183 & 79.2 & 43.2 & 48.2 & 63.4 & 54.7 \\
& ILLUME~\cite{illume}            & 7B           & 1445 & -    & 75.1 & 38.2 & 37.0 & -    & -    \\
& VILA-U$^{**}_{256}$~\cite{vila-u}  & 7B           & 1336 & -    & 66.6 & 32.2 & 27.7 & -    & 22.0 \\
& Chameleon$^{**}$~\cite{chameleon}       & 7B           & -    & -    & 35.7 & 28.4 & 8.3  & -    & 0.0  \\
& Janus-Pro~\cite{janus-pro}         & 7B           & 1567 & -    & 79.2 & 41.0 & 50.0 & -    & -    \\
& MetaQuery-XL$^\dagger$~\cite{metaqueries} & 7B      & \underline{1685} & -    & 83.5 & \textbf{58.6} & 66.6 & -    & -    \\
& LlamaFusion$^{**}$~\cite{lmfusion}     & 8B           & 1604 & -    & 72.1 & 41.7 & -    & -    & -    \\
& MetaMorph~\cite{metamorph}         & 8B           & -    & -    & 75.2 & 41.8 & -    & -    & 48.3 \\
& SEED-X~\cite{seed-x}            & 13B          & 1457 & -    & 70.1 & 35.6 & 43.0 & -    & -    \\
& TokenFlow-XL~\cite{tokenflow}      & 13B          & 1546 & -    & 68.9 & 38.7 & 40.7 & -    & -    \\
& MUSE-VL~\cite{muse-vl}           & 32B          & -    & -    & 81.8 & 50.1 & -    & 55.9 & -    \\
& BAGEL~\cite{bagel}         & 7B MoT       & \textbf{1687} & \textbf{2388} & \textbf{85.0} & 53.7$^{*}$ & \textbf{67.2} & \underline{73.1} & \textbf{69.3} \\

\rowcolor{blue!20}
& \textbf{ChatUMM (Ours)} & 7B MoT & \underline{1685} & \underline{2382} & \underline{84.6} & 53.8$^{*}$ & 66.4 & \textbf{74.7} & \underline{68.7} \\

\bottomrule
\end{tabular}
}
\vspace{1em}
\end{table*}

\begin{figure}[t!]
    \centering
    \includegraphics[width=\textwidth]{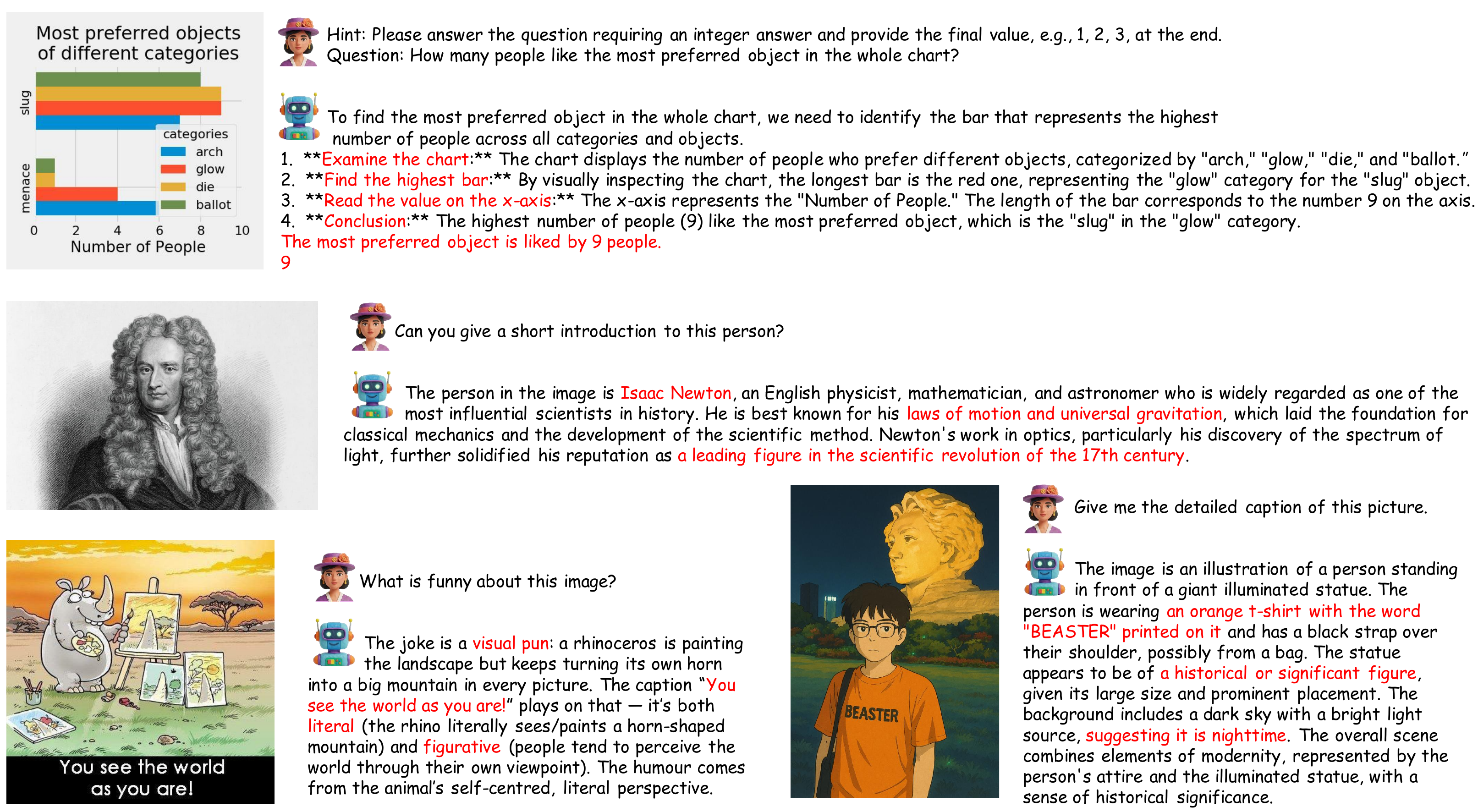}
    \caption{\textbf{Qualitative examples of multimodal understanding capabilities.} ChatUMM demonstrates versatility in diverse tasks, including chart interpretation and reasoning, celebrity recognition with knowledge retrieval, explanation of visual humor, and detailed image captioning.}
    \vspace{1em}
    \label{figures/und_vis}
\end{figure}

We evaluate ChatUMM’s core visual understanding capabilities on a wide range of benchmarks, including MME, MMBench, MMMU, MM-Vet, MathVista, and MMVP. The quantitative results are presented in~\cref{tab:model_performance_und}. ChatUMM delivers top-tier performance among open-source unified models~\cite{janus-pro, metaqueries, bagel}, achieving \textbf{84.6} on MMBench, \textbf{53.8} on MMMU, \textbf{66.4} on MM-Vet, and \textbf{74.7} on MathVista. Crucially, our model performs strictly on par with, and in several cases surpasses, the powerful BAGEL (7B MoT) baseline~\cite{bagel}. ChatUMM maintains comparable scores on perception-heavy benchmarks (e.g., MME-P, MMVP) while exhibiting a slight improvement on reasoning-intensive benchmarks such as MathVista (\textbf{+1.6}). As shown in~\cref{figures/und_vis}, these capabilities are further supported by qualitative examples. The model successfully handles diverse tasks, ranging from multi-step chart reasoning and celebrity recognition (e.g., Isaac Newton) to explaining visual humor and generating detailed image captions, demonstrating its robust visual understanding.

\subsection{Image Generation}

\begin{table*}[t!]
\vspace{-1em}
\centering
\caption{\textbf{Evaluation of text-to-image generation ability on GenEval benchmark.} `Gen. Only' stands for an image generation model, and `Unified' denotes a model that has both understanding and generation capabilities. $\dagger$ refers to the methods using the LLM rewriter.}
\vspace{-0.5em}
\label{tab:model_performance_gen}
\resizebox{\textwidth}{!}{
\begin{tabular}{llccccccc}
\toprule
\textbf{Type} & \textbf{Model} & \textbf{Single Obj.} & \textbf{Two Obj.} & \textbf{Counting} & \textbf{Colors} & \textbf{Position} & \textbf{Color Attri.} & \textbf{Overall}$\uparrow$ \\
\midrule
\multirow{8}{*}{\rotatebox{90}{Gen. Only}}
& PixArt-$\alpha$~\cite{pixart} & 0.98 & 0.50 & 0.44 & 0.80 & 0.08 & 0.07 & 0.48 \\
& SDv2.1~\cite{sd} & 0.98 & 0.51 & 0.44 & 0.85 & 0.07 & 0.17 & 0.50 \\
& DALL-E 2~\cite{dalle2} & 0.94 & 0.66 & 0.49 & 0.77 & 0.10 & 0.19 & 0.52 \\
& Emu3-Gen~\cite{emu3} & 0.98 & 0.71 & 0.34 & 0.81 & 0.17 & 0.21 & 0.54 \\
& SDXL~\cite{sdxl} & 0.98 & 0.74 & 0.39 & 0.85 & 0.15 & 0.23 & 0.55 \\
& DALL-E 3~\cite{dalle3} & 0.96 & 0.87 & 0.47 & 0.83 & 0.43 & 0.45 & 0.67 \\
& SD3-Medium~\cite{sd3} & 0.99 & 0.94 & 0.72 & 0.89 & 0.33 & 0.60 & 0.74 \\
& FLUX.1-dev$^\dagger$~\cite{flux} & 0.98 & 0.93 & 0.75 & 0.93 & 0.68 & 0.65 & 0.82 \\
\midrule
\multirow{12}{*}{\rotatebox{90}{Unified}}
& Chameleon~\cite{chameleon} & - & - & - & - & - & - & 0.39 \\
& LWM~\cite{lwm} & 0.93 & 0.41 & 0.46 & 0.79 & 0.09 & 0.15 & 0.47 \\
& SEED-X~\cite{seed-x} & 0.97 & 0.58 & 0.26 & 0.80 & 0.19 & 0.14 & 0.49 \\
& TokenFlow-XL~\cite{tokenflow} & 0.95 & 0.60 & 0.41 & 0.81 & 0.16 & 0.24 & 0.55 \\
& ILLUME~\cite{illume} & 0.99 & 0.86 & 0.45 & 0.71 & 0.39 & 0.28 & 0.61 \\
& Janus~\cite{janus} & 0.97 & 0.68 & 0.30 & 0.84 & 0.46 & 0.42 & 0.61 \\
& Transfusion~\cite{transfusion} & - & - & - & - & - & - & 0.63 \\
& Emu3-Gen$^\dagger$~\cite{emu3} & 0.99 & 0.81 & 0.42 & 0.80 & 0.49 & 0.45 & 0.66 \\
& Show-o~\cite{show-o} & 0.98 & 0.80 & 0.66 & 0.84 & 0.31 & 0.50 & 0.68 \\
& Janus-Pro-7B~\cite{janus-pro} & 0.99 & 0.89 & 0.59 & 0.90 & 0.79 & 0.66 & 0.80 \\
& MetaQuery-XL$^\dagger$~\cite{metaqueries} & - & - & - & - & - & - & 0.80 \\
& BAGEL$^\dagger$~\cite{bagel} & 0.98 & 0.95 & 0.84 & 0.95 & 0.78 & 0.77 & \textbf{0.88} \\

\rowcolor{blue!20}
& \textbf{ChatUMM (Ours)$^\dagger$} & 0.99 & 0.96 & 0.79 & 0.92 & 0.69 & 0.77 & \underline{0.85} \\

\bottomrule
\end{tabular}
}
\end{table*}

\begin{figure}[t!]
    \centering
    \includegraphics[width=\textwidth]{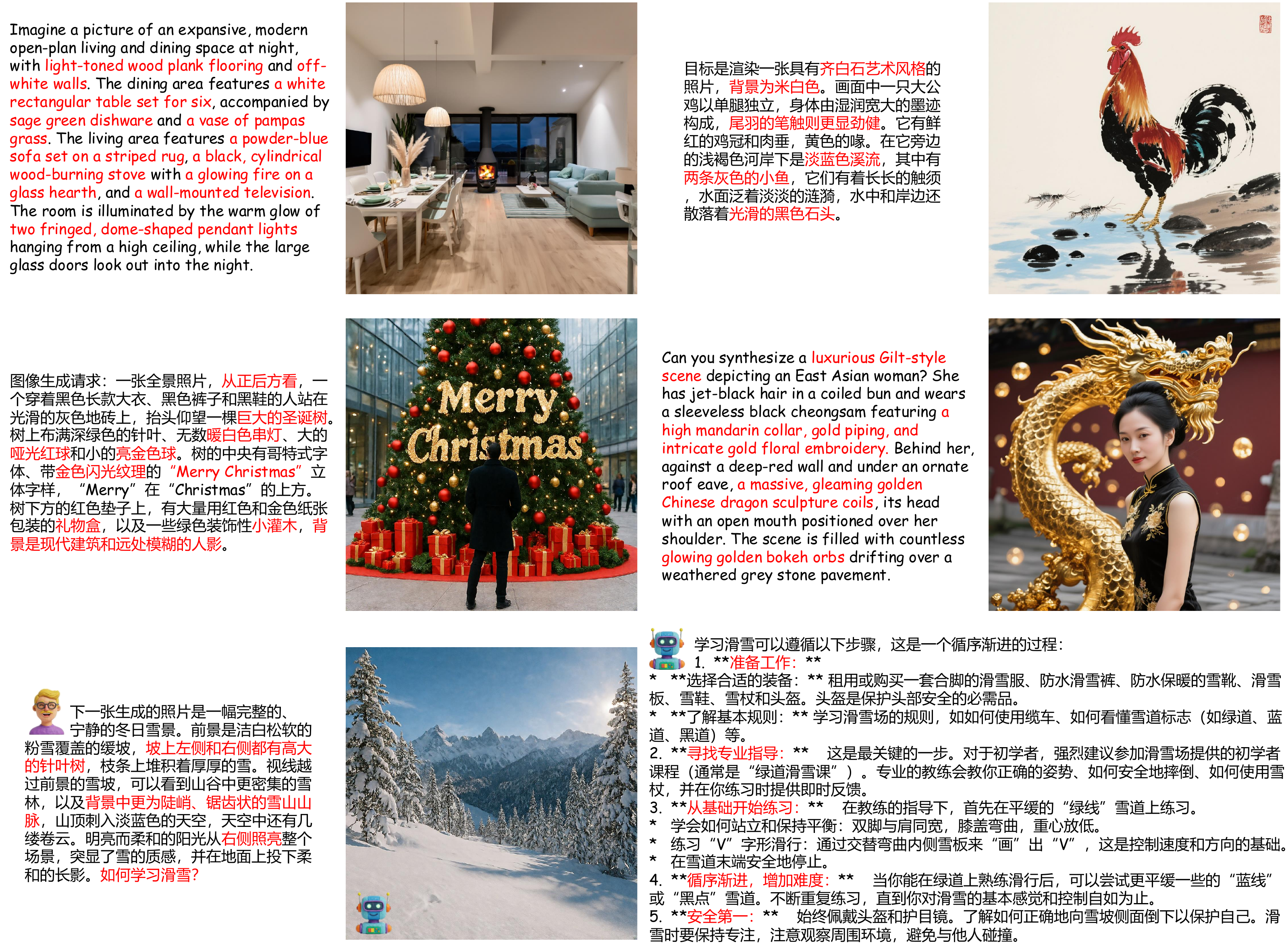}
    \vspace{-1.5em}
    \caption{\textbf{Qualitative examples of text-to-image.} ChatUMM excels at generating high-fidelity images across various styles, such as photorealistic interiors, traditional Chinese paintings, and Gilt-style portraits. It accurately renders specific text elements (e.g., ``Merry Christmas'') and lighting details described in long, detailed prompts. Notably, the bottom example highlights the interleaved generation capability: the model generates a snowy landscape and provides a step-by-step skiing tutorial in response to the user's request.}
    \vspace{-2em}
    \label{figures/t2m_vis}
\end{figure}

To assess text-to-image generation quality, we evaluate ChatUMM on the GenEval benchmark, which systematically measures the alignment between generated images and textual prompts covering diverse attributes like object count, spatial relationships, and color accuracy. 
As shown in~\cref{tab:model_performance_gen}, ChatUMM achieves an overall score of \textbf{0.85}, substantially outperforming unified baselines like Janus-Pro-7B (\textbf{+0.05}) and MetaQuery-XL (\textbf{+0.05}). 
Notably, it also exceeds the performance of specialist generation models, surpassing SD3-Medium by \textbf{0.11} and even FLUX.1-dev by \textbf{0.03}.
We show qualitative results in~\cref{figures/t2m_vis}. 
The model accurately renders specific text elements and handles diverse styles, ranging from photorealistic interiors and traditional Chinese paintings to Gilt-style portraits. 
Furthermore, the bottom example highlights the interleaved generation capability, where the model generates a snowy landscape followed by a step-by-step skiing tutorial. 
These results provide strong evidence of the superior image generation capabilities of ChatUMM.

\subsection{Image Editing}

\begin{table*}[t!]
\vspace{-1em}
\centering
\caption{\textbf{Comparison on GEdit-Bench.} All metrics are reported as higher-is-better ($\uparrow$). $G\_SC$, $G\_PQ$, and $G\_O$ refer to the metrics evaluated by GPT-4.1.}
\label{tab:model_performance_edit}
\resizebox{\textwidth}{!}{
\begin{tabular}{ll|ccc|ccc}
\toprule
\textbf{Type} & \textbf{Model} & \multicolumn{3}{c|}{\textbf{GEdit-Bench-EN (Full set)}$\uparrow$} & \multicolumn{3}{c}{\textbf{GEdit-Bench-CN (Full set)}$\uparrow$} \\
\cmidrule(lr){3-5} \cmidrule(lr){6-8}
& & \textbf{G\_SC} & \textbf{G\_PQ} & \textbf{G\_O} & \textbf{G\_SC} & \textbf{G\_PQ} & \textbf{G\_O} \\
\midrule
\textit{Private} & Gemini 2.0~\cite{gemini-2.0-flash-exp} & 6.73 & 6.61 & 6.32 & 5.43 & 6.78 & 5.36 \\
& GPT-4o~\cite{gpt4o} & \textbf{7.85} & \textbf{7.62} & \textbf{7.53} & \textbf{7.67} & \textbf{7.56} & \textbf{7.30} \\
\midrule
\textit{Open-source} & Instruct-Pix2Pix~\cite{instructpix2pix} & 3.58 & 5.49 & 3.68 & - & - & - \\
& MagicBrush~\cite{magicbrush} & 4.68 & 5.66 & 4.52 & - & - & - \\
& AnyEdit~\cite{anyedit} & 3.18 & 5.82 & 3.21 & - & - & - \\
& OmniGen~\cite{omnigen} & 5.96 & 5.89 & 5.06 & - & - & - \\
& Step1X-Edit~\cite{step1x} & 7.09 & 6.76 & 6.70 & 7.20 & 6.87 & \underline{6.86} \\
& BAGEL~\cite{bagel} & 7.36 & 6.83 & 6.52 & 7.34 & 6.85 & 6.50 \\

\rowcolor{blue!20}
& \textbf{ChatUMM (Ours)} & \underline{7.69} & \underline{7.21} & \underline{6.95} & \underline{7.40} & \underline{7.24} & 6.73 \\

\bottomrule
\end{tabular}
}
\end{table*}

\begin{figure}[t!]
    \centering
    \includegraphics[width=\textwidth]{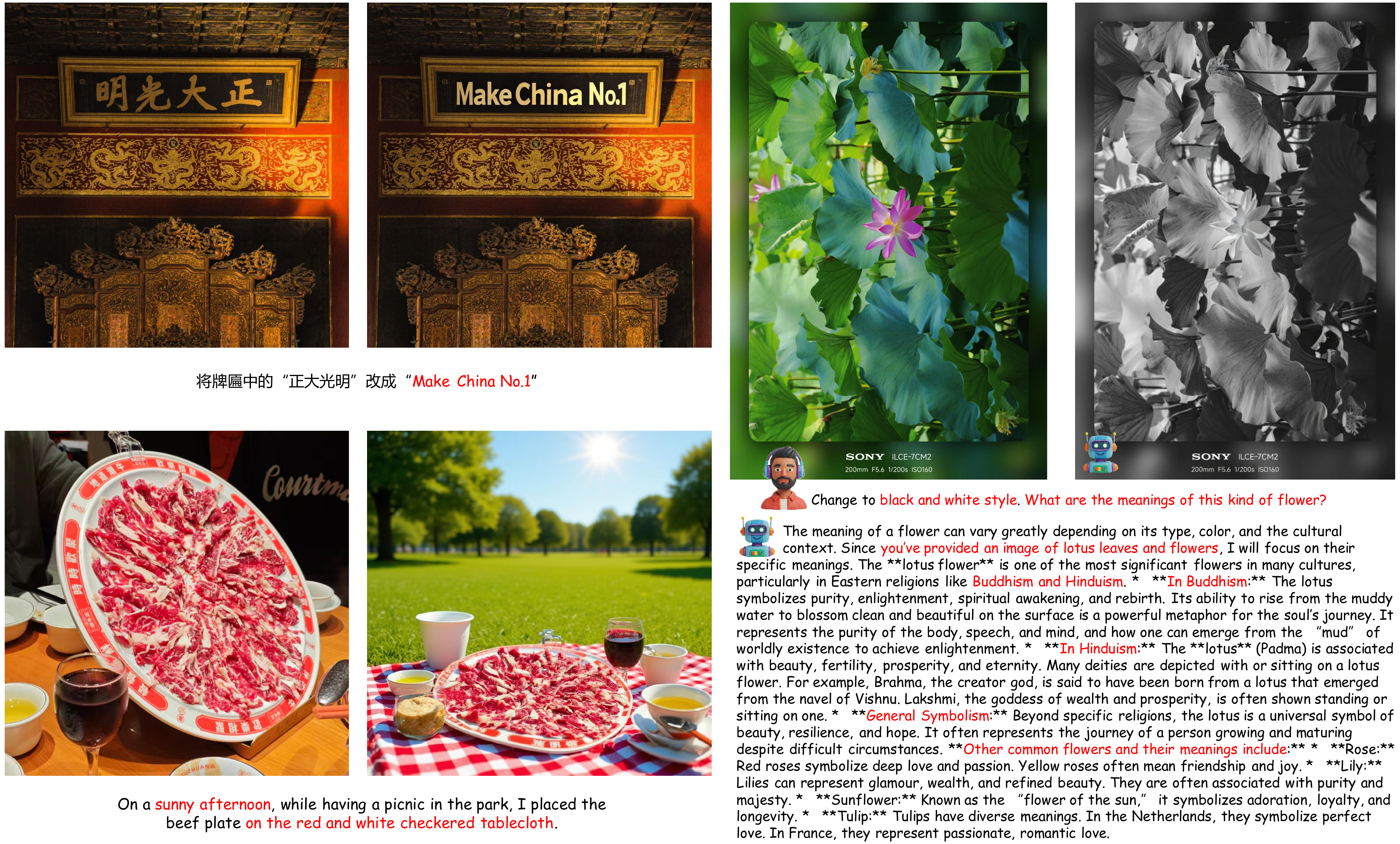}
    \vspace{-1.5em}
    \caption{\textbf{Qualitative examples of instruction-guided image editing.} ChatUMM performs precise text replacement within images and integrates objects into scenes. Notably, the right example showcases the model's interleaved generation capability: it executes a style transfer while simultaneously explaining the cultural meaning of the flower in the provided image.}
    \vspace{-1em}
    \label{figures/edit_vis}
\end{figure}

\begin{figure}[t!]
    \vspace{-1.5em}
    \centering
    \includegraphics[width=\textwidth]{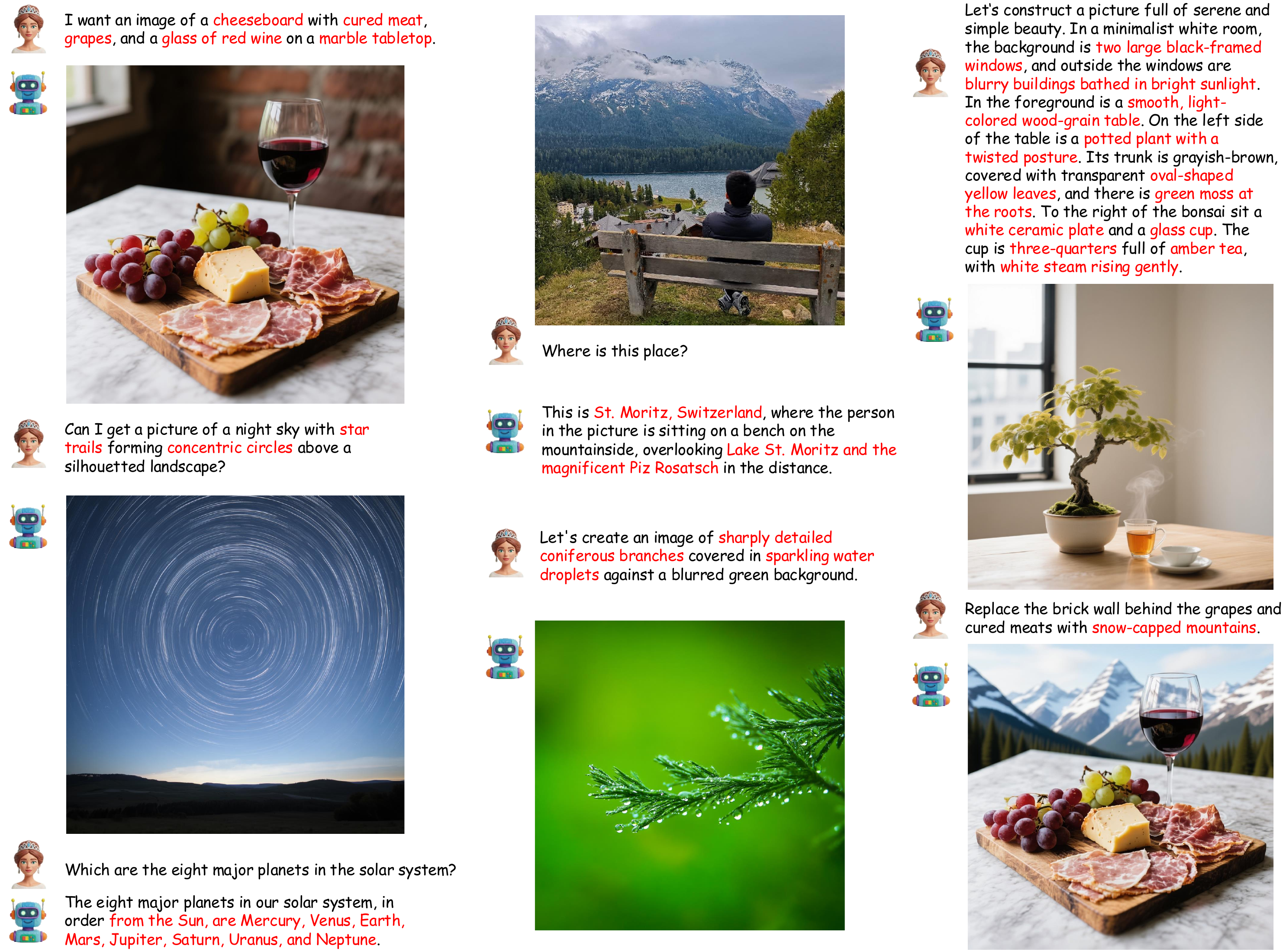}
    \vspace{-1.5em}
    \caption{\textbf{Example of robust multi-turn capabilities.} ChatUMM seamlessly interleaves diverse tasks, including image generation, visual reasoning, and factual Q\&A, within a continuous conversational flow. The final turn highlights its long-range image editing capability: the model executes the instruction via accurate retrieval of the initial image across multiple ``distractor'' turns.}
    \vspace{-1.5em}
    \label{figures/edit_long_vis}
\end{figure}

We evaluate the instruction-guided image editing capabilities on the GEdit-Bench benchmark.
As detailed in~\cref{tab:model_performance_edit}, ChatUMM establishes a new state-of-the-art for open-source unified models, outperforming the baseline BAGEL on both English and Chinese sets.
Specifically, on the English split, it achieves an overall score ($G\_O$) of \textbf{6.95}, exceeding BAGEL by \textbf{0.43} and the proprietary Gemini 2.0 by \textbf{0.63}.
Crucially, it also surpasses the editing specialist Step1X-Edit by \textbf{0.25} in this setting.
Qualitative examples in~\cref{figures/edit_vis} demonstrate precise object integration and text replacement.
Notably, it exhibits interleaved generation capabilities, such as executing style transfer while simultaneously interpreting visual content.
This confirms the effectiveness of ChatUMM in handling various image editing tasks through robust instruction following.

\begin{figure}[h!]
    \vspace{-2em}
    \centering
    \includegraphics[width=0.95\textwidth]{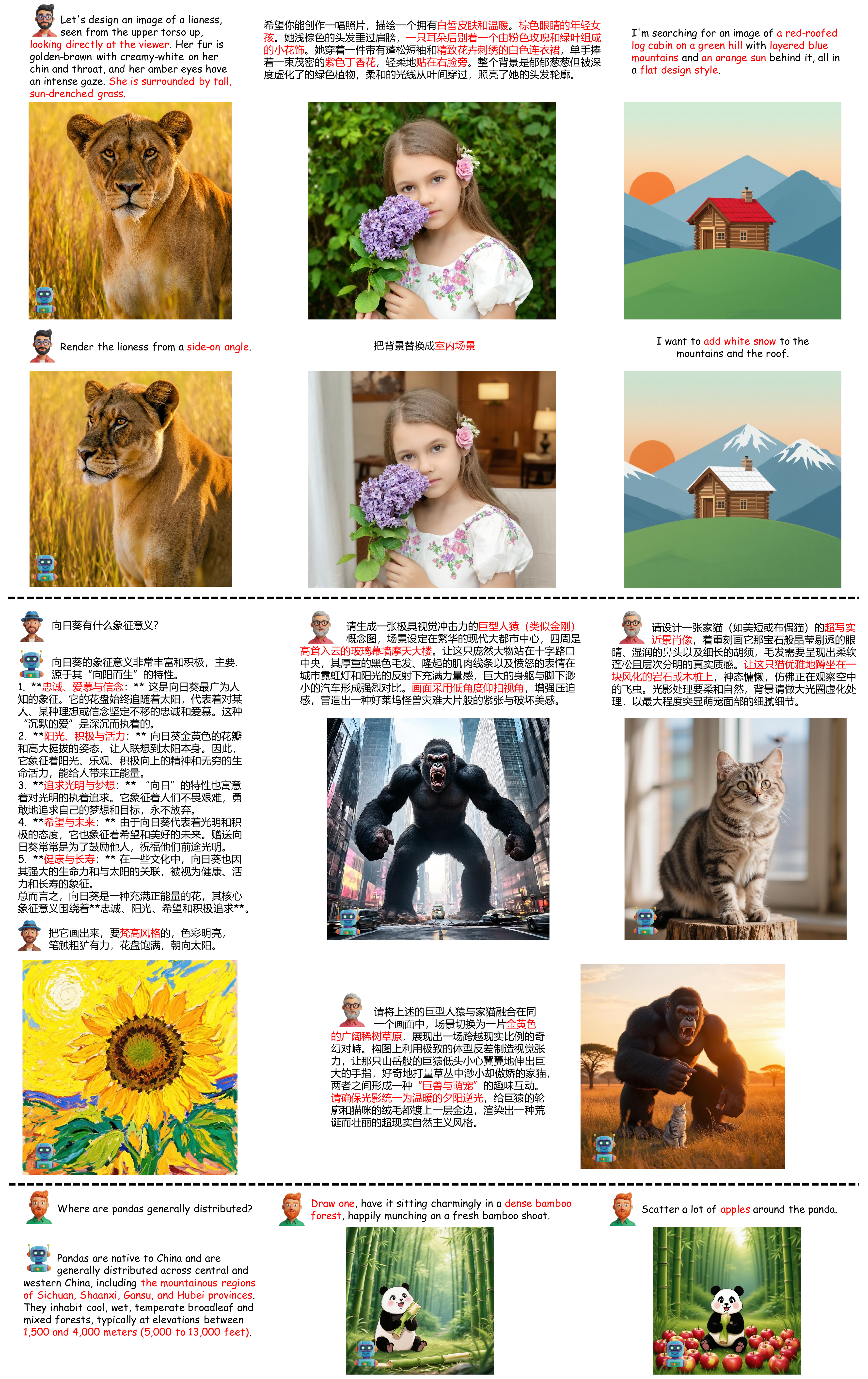}
    \caption{\textbf{Basic multi-turn interactions.} \textbf{Top:} The model ensures subject consistency during editing. \textbf{Middle:} Context-aware generation from text history (left) and merging prior subjects into a single scene (right). \textbf{Bottom:} An interleaved sequence of knowledge retrieval, generation, and editing.}
    \vspace{-2em}
    \label{figures/simple_mt}
\end{figure}

\begin{figure}[h!]
    \vspace{-1em}
    \centering
    \includegraphics[width=\textwidth]{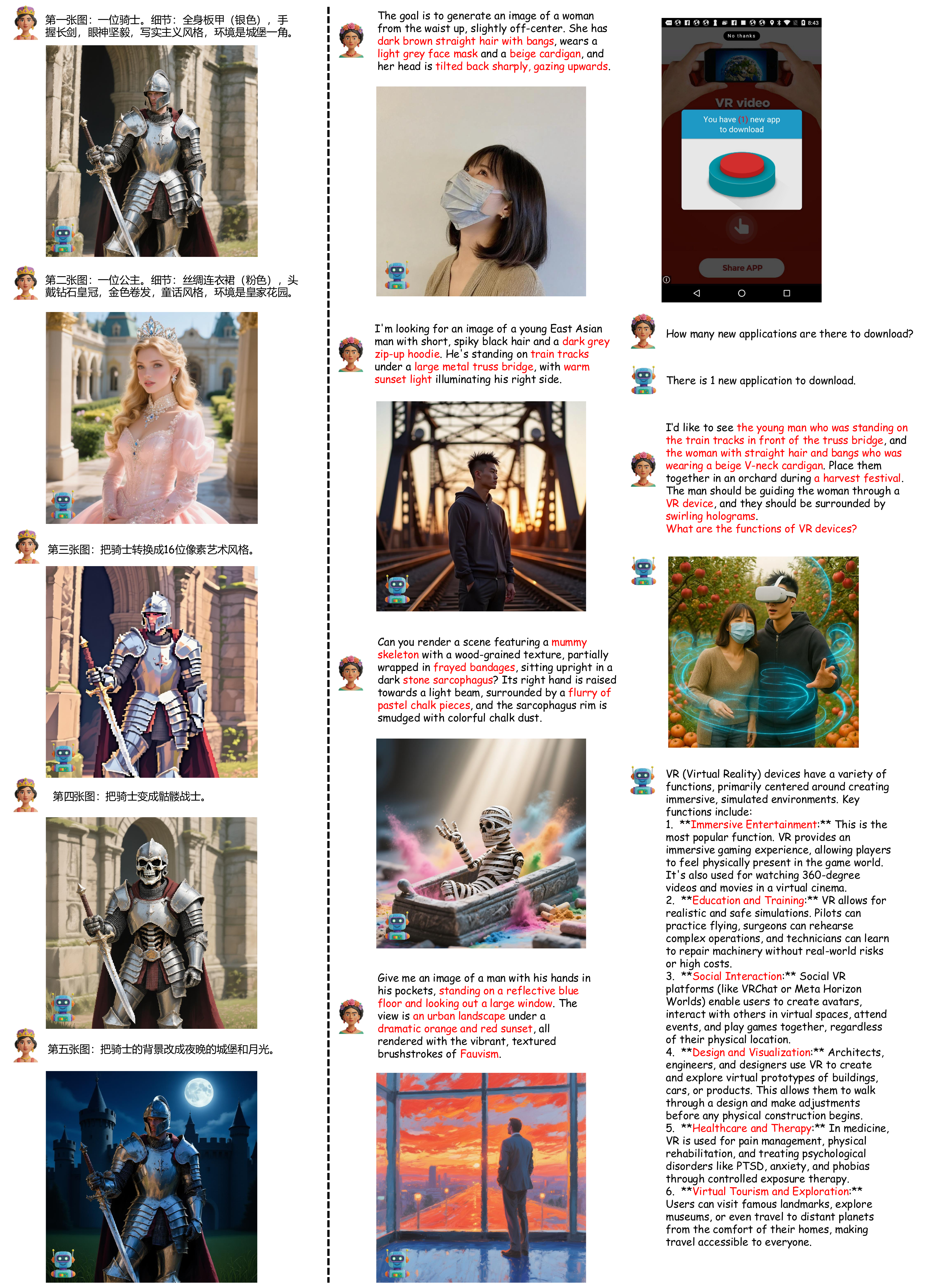}
    \caption{\textbf{Advanced multi-turn dialogues.} \textbf{Left:} Targeted subject editing. The model correctly edits the ``Knight'' from the first turn, effectively bypassing the unrelated ``Princess'' generated in the second turn to perform history-dependent editing. \textbf{Right:} Long-range subject-driven generation. The model combines two characters from the initial turns, effectively filtering out a sequence of “distractor” turns (e.g., Mummy, Fauvism painting) to execute the final composition.}
    \vspace{-1em}
    \label{figures/hard_mt}
\end{figure}

\vspace{-0.75em}

\section{Conclusion}

\vspace{-0.75em}

In this work, we present ChatUMM, a conversational unified model that breaks the single-turn paradigm to enable interleaved multimodal generation. Through our interleaved multi-turn training strategy and systematic data synthesis pipeline, ChatUMM demonstrates robust context tracking and accurate intent disambiguation. Extensive evaluations confirm that our model achieves state-of-the-art performance in visual understanding and editing, maintains high fidelity in text-to-image generation, and exhibits superior robustness in complex dialogue scenarios.

Looking forward, we identify key directions to further advance conversational unified models. First, while ChatUMM validates the feasibility of the unified architecture for sustained, multi-turn interactions, closing the gap with commercial agent-based systems (e.g., GPT-4o) will require scaling model parameters and integrating more complex reasoning data. Second, to address the computational costs associated with the dual-encoder design, developing a ``unified tokenizer'' that effectively captures high-level semantics and preserves low-level visual details without compromising inference efficiency remains a critical direction for future research. Ultimately, ChatUMM serves as a foundational step, offering a strong baseline for the community to explore fluid, end-to-end multimodal interactions beyond the limitations of current tool-use frameworks.

\newpage
\bibliography{colm2024_conference}
\bibliographystyle{colm2024_conference}

\end{document}